\documentclass[conference]{IEEEtran}
\usepackage{cite}
\usepackage{graphicx}
\usepackage{latexsym}
\usepackage{amssymb}
\usepackage{url}
\usepackage[fleqn]{amsmath}
\usepackage[varg]{txfonts}
\usepackage{algpseudocode}
\usepackage{refcount}
\usepackage{footnote}
\makesavenoteenv{tabular}
\makesavenoteenv{table}

\ifCLASSINFOpdf
\else
\fi
\begin{document}
\title{Gated Recurrent Neural Tensor Network}


\author{\IEEEauthorblockN{Andros Tjandra\IEEEauthorrefmark{1},
Sakriani Sakti\IEEEauthorrefmark{2},
Ruli Manurung\IEEEauthorrefmark{1}, 
Mirna Adriani\IEEEauthorrefmark{1} and
Satoshi Nakamura\IEEEauthorrefmark{2}}
\IEEEauthorblockA{\IEEEauthorrefmark{1} Faculty of Computer Science, Universitas Indonesia, Indonesia\\
Email: andros.tjandra@gmail.com, ruli.manurung@cs.ui.ac.id, mirna@cs.ui.ac.id}
\IEEEauthorblockA{\IEEEauthorrefmark{2}Graduate School of Information Science, Nara Institute of Science and Technology, Japan\\
Email : ssakti@is.naist.jp, s-nakamura@is.naist.jp}
}

%Telephone: (800) 555--1212, Fax: (888) 555--1212}


\maketitle

\begin{abstract}
Recurrent Neural Networks (RNNs), which are a powerful scheme for modeling temporal and sequential data need to capture long-term dependencies on datasets and represent them in hidden layers with a powerful model to capture more information from inputs. For modeling long-term dependencies in a dataset, the gating mechanism concept can help RNNs remember and forget previous information. Representing the hidden layers of an RNN with more expressive operations (i.e., tensor products) helps it learn a more complex relationship between the current input and the previous hidden layer information. These ideas can generally improve RNN performances. In this paper, we proposed a novel RNN architecture that combine the concepts of gating mechanism and the tensor product into a single model. By combining these two concepts into a single RNN, our proposed models learn long-term dependencies by modeling with gating units and obtain more expressive and direct interaction between input and hidden layers using a tensor product on 3-dimensional array (tensor) weight parameters. We use Long Short Term Memory (LSTM) RNN and Gated Recurrent Unit (GRU) RNN and combine them with a tensor product inside their formulations. Our proposed RNNs, which are called a Long-Short Term Memory Recurrent Neural Tensor Network (LSTMRNTN) and Gated Recurrent Unit Recurrent Neural Tensor Network (GRURNTN), are made by combining the LSTM and GRU RNN models with the tensor product. We conducted experiments with our proposed models on word-level and character-level language modeling tasks and revealed that our proposed models significantly improved their performance compared to our baseline models.
\end{abstract}


% \fi
\IEEEpeerreviewmaketitle

\section{Introduction}\label{intro}
Modeling temporal and sequential data, which is crucial in machine learning, can be applied in many areas, such as speech and natural language processing. Deep neural networks (DNNs) have garnered interest from many researchers after being successfully applied in image classification \cite{krizhevsky2012imagenet} and speech recognition \cite{mohamed2012acoustic}. Another type of neural network, called a recurrent neural network (RNN) is also widely used for speech recognition \cite{graves2013speech}, machine translation \cite{sutskever2014sequence, cho2014learning} and language modeling \cite{mikolov2010recurrent, sundermeyer2012lstm}. RNNs have achieved many state-of-the-art results. Compared to DNNs, they have extra parameters for modeling the relationships of previous or future hidden states with current input where the RNN parameters are shared for each input time-step.

Generally, RNNs can be separated by a simple RNN without gating units, such as the Elman RNN \cite{elman1990finding}, the Jordan RNN \cite{jordan1997serial}, and such advanced RNNs with gating units as the Long-Short Term Memory (LSTM) RNN \cite{hochreiter1997long} and the Gated Recurrent Unit (GRU) RNN\cite{cho2014learning}. A simple RNN usually adequate to model some dataset and a task with short-term dependencies like slot filling for spoken language understanding \cite{mesnil2013investigation}. However, for more difficult tasks like language modeling and machine translation where most predictions need longer information and a historical context from each sentence, gating units are needed to achieve good performance. With gating units for blocking and passing information from previous or future hidden layer, we can learn long-term information and recursively backpropagate the error from our prediction without suffering from vanishing or exploding gradient problems \cite{hochreiter1997long}. In spite of this situation, the concept of gating mechanism does not provide an RNN with a more powerful way to model the relation between the current input and previous hidden layer representations.

Most interactions inside RNNs between current input and previous (or future) hidden states are represented using linear projection and addition and are transformed by the nonlinear activation function. The transition is shallow because no intermediate hidden layers exist for projecting the hidden states \cite{pascanu2013construct}. To get a more powerful representation on the hidden layer, Pascanu et al.\cite{pascanu2013construct} modified RNNs with an additional nonlinear layer from input to the hidden layer transition, hidden to hidden layer transition and also hidden to output layer transition. Socher et al.\cite{socher2013recursive, socher2013reasoning} proposed another approach using a tensor product for calculating output vectors given two input vectors. They modified a Recursive Neural Network (RecNN) to overcome those limitations using more direct interaction between two input layers. This architecture is called a Recursive Neural Tensor Network (RecNTN), which uses a tensor product between child input vectors to represent the parent vector representation. By adding the tensor product operation to calculate their parent vector, RecNTN significantly improves the performance of sentiment analysis and reasoning on entity relations tasks compared to standard RecNN architecture. However, those models struggle to learn long-term dependencies because the do not utilize the concept of gating mechanism.  

In this paper, we proposed a new RNN architecture that combine the gating mechanism and tensor product concepts to incorporate both advantages in a single architecture. Using the concept of such gating mechanisms as LSTMRNN and GRURNN, our proposed architecture can learn temporal and sequential data with longer dependencies between each input time-step than simple RNNs without gating units and combine the gating units with tensor products to represent the hidden layer with more powerful operation and direct interaction. Hidden states are generated by the interaction between current input and previous (or future) hidden states using a tensor product and a non-linear activation function allows more expressive model representation. We describe two different models based on LSTMRNN and GRURNN. LSTMRNTN is our proposed model for the combination between a LSTM unit with a tensor product inside its cell equation and GRURNTN is our name for a GRU unit with a tensor product inside its candidate hidden layer equation.

In Section \ref{sec:background}, we provide some background information related to our research. In Section \ref{sec:proposed}, we describe our proposed RNN architecture in detail. We evaluate our proposed RNN architecture on word-level and character-level language modeling tasks and reported the result in Section \ref{sec:expr}. We present related works in Section \ref{sec:rel}. Section \ref{sec:conclusion} summarizes our paper and provides some possible future improvements.

\section{Background} \label{sec:background}
\subsection{Recurrent Neural Network} \label{sec:rnn}
A Recurrent Neural Network (RNN) is one kind of neural network architecture for modeling sequential and temporal dependencies \cite{graves2013speech}. Typically, we have input sequence $\mathbf{x}=(x_1,...,x_{T})$ and calculate hidden vector sequence $\mathbf{h}=(h_1,...,h_{T})$ and output vector sequence $\mathbf{y}=(y_1,...,y_T)$ with RNNs. A standard RNN at time $t$-th is usually formulated as:
\begin{eqnarray}
h_t &=& f(x_t W_{xh} + h_{t-1} W_{hh} + b_h) \\
y_t &=& g(h_t W_{hy} + b_y). 
\end{eqnarray} where $W_{xh}$ represents the input layer to the hidden layer weight matrix, $W_{hh}$ represents hidden to hidden layer weight matrix, $W_{hy}$ represents the hidden to the output weight matrix, $b_h$ and $b_y$ represent bias vectors for the hidden and output layers. $f(\cdot)$ and $g(\cdot)$ are nonlinear activation functions such as sigmoid or tanh.

\begin{figure}[h]
    \centering
    \includegraphics[width=3.5cm]{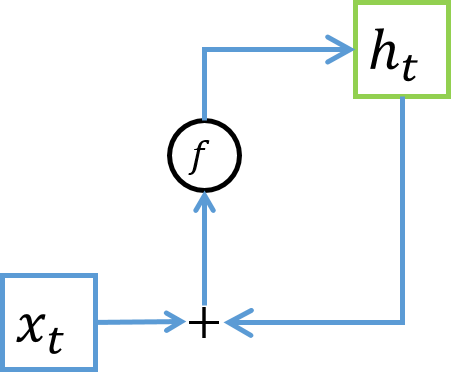}
    \caption{Recurrent Neural Network}
    \label{fig:rnn}
\end{figure}

\subsection{Gated Recurrent Neural Network} \label{sec:gatedrnn}
Simple RNNs are hard to train to capture long-term dependencies from long sequential datasets because the gradient can easily explode or vanish \cite{bengio1994learning, hochreiter2001gradient}. Because the gradient (usually) vanishes after several steps, optimizing a simple RNN is more complicated than standard neural networks. To overcome the disadvantages of simple RNNs, several researches have been done. Instead of using a first-order optimization method, one approach optimized the RNN using a second-order Hessian Free optimization \cite{martens2011learning}. Another approach, which addressed the vanishing and exploding gradient problem, modified the RNN architecture with additional parameters to control the information flow from previous hidden layers using the gating mechanism concept \cite{hochreiter1997long}. A gated RNN is a special recurrent neural network architecture that overcomes this weakness of a simple RNN by introducing gating units. There are variants from RNN with gating units, such as Long Short Term Memory (LSTM) RNN and Gated Recurrent Unit (GRU) RNN. In the following sections, we explain both LSTMRNN and GRURNN in more detail.

\subsubsection{Long Short Term Memory RNN} \label{sec:lstm}
A Long Short Term Memory (LSTM) \cite{hochreiter1997long} is a gated RNN with three gating layers and memory cells. The gating layers are used by the LSTM to control the existing memory by retaining the useful information and forgetting the unrelated information. Memory cells are used for storing the information across time. The LSTM hidden layer at time $t$ is defined by the following equations \cite{graves2013hybrid}:
\begin{eqnarray}
i_t &=& \sigma(x_t W_{xi} + h_{t-1} W_{hi} + c_{t-1} W_{ci} + b_i) \\
f_t &=& \sigma(x_t W_{xf} + h_{t-1} W_{hf} + c_{t-1} W_{cf} + b_f) \\
c_t &=& f_t \odot c_{t-1} + i_t \odot \tanh(x_t W_{xc} + h_{t-1} W_{hc} + b_c) \\
o_t &=& \sigma(x_t W_{xo} + h_{t-1} W_{ho} + c_t W_{co} + b_o) \\
h_t &=& o_t \odot \tanh(c_t)
\end{eqnarray}
where $\sigma(\cdot)$ is sigmoid activation function and $i_t, f_t, o_t$ and $c_t$ are respectively the input gates, the forget gates, the output gates and the memory cells at time-step $t$. The input gates keep the candidate memory cell values that are useful for memory cell computation, and the forget gates keep the previous memory cell values that are useful for calculating the current memory cell. The output gates filter which the memory cell values that are useful for the output or next hidden layer input. 

\begin{figure}[h]
    \centering
    \includegraphics[width=6.5cm]{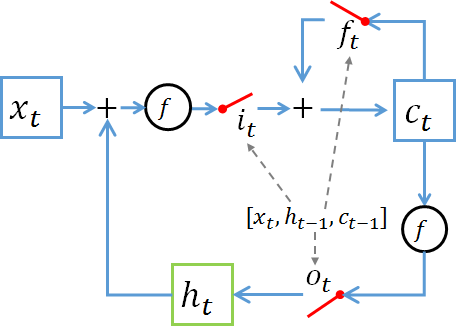}
    \caption{Long Short Term Memory Unit.}
    \label{fig:lstmrnn}
\end{figure}

\subsubsection{Gated Recurrent Unit RNN} \label{sec:gru}
A Gated Recurrent Unit (GRU) \cite{cho2014learning} is a gated RNN with similar properties to a LSTM. However, there are several differences: a GRU does not have separated memory cells \cite{chung2014empirical}, and instead of three gating layers, it only has two gating layers: reset gates and update gates.
The GRU hidden layer at time $t$ is defined by the following equations \cite{cho2014learning}:
\begin{eqnarray}
r_t &=& \sigma(x_t W_{xr} + h_{t-1} W_{hr} + b_r)\\
z_t &=& \sigma(x_t W_{xz} + h_{t-1} W_{hz} + b_r)\\
\tilde{h_t} &=& f(x_t W_{xh} + (r_t \odot h_{t-1}) W_{hh} + b_h)\\
h_t &=& (1 - z_t) \odot h_{t-1} + z_t \odot \tilde{h_t}
\end{eqnarray}
where $\sigma(\cdot)$ is a sigmoid activation function, $r_t, z_t$ are reset and update gates, $\tilde{h_t}$ is the candidate hidden layer values and $h_t$ is the hidden layer values at time-$t$. The reset gates determine which previous hidden layer value is useful for generating the current candidate hidden layer. The update gates keeps the previous hidden layer values or replaced by new candidate hidden layer values. 
In spite of having one fewer gating layer, the GRU can match LSTM's performance and its convergence speed convergence sometimes outperformed LSTM \cite{chung2014empirical}.
\begin{figure}[h]
    \centering
    \includegraphics[width=6.5cm]{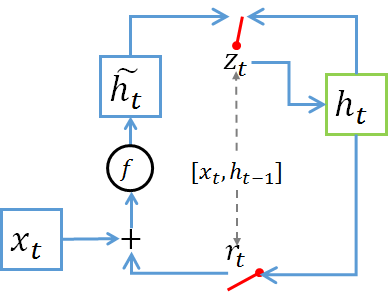}
    \caption{Gated Recurrent Unit}
    \label{fig:grurnn}
\end{figure}

\subsection{Recursive Neural Tensor Network} \label{sec:recntn}
A Recursive Neural Tensor Network (RecNTN) is a variant of a Recursive Neural Network (RecNN) for modeling input data with variable length properties and tree structure dependencies between input features \cite{socher2011parsing}. To compute the input representation with RecNN, the input must be parsed into a binary tree where each leaf node represents input data. Then, the parent vectors are computed in a bottom-up fashion, following the above computed tree structure whose information can be built using external computation tools (i.e., syntactic parser) or some heuristic from our dataset observations.
\begin{figure}[h]
    \centering
    \includegraphics[width=5.5cm]{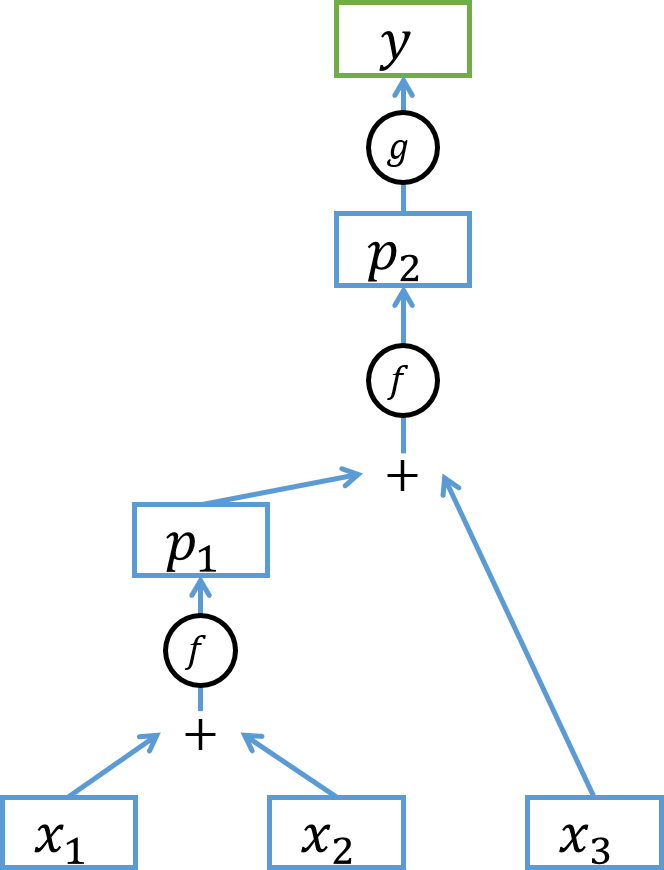}
    \caption{Computation for parent values $p_1$ and $p_2$ was done in a bottom-up fashion. Visible node leaves $x_1, x_2$, and $x_3$ are processed based on the given binary tree structure.}
    \label{fig:recnn}
\end{figure}
Given Fig. \ref{fig:recnn}, $p_1$, $p_2$ and $y$ was defined by:
\begin{eqnarray}
\label{eq:p1} p_1 &=& f\left( \begin{bmatrix} x_1 & x_2 \end{bmatrix} W + b \right) \\ 
\label{eq:p2} p_2 &=& f\left( \begin{bmatrix} p_1 & x_3 \end{bmatrix} W + b \right)  \\
y &=& g\left( p_2 W_y + b_y \right)
\end{eqnarray} where $f(\cdot)$ is nonlinear activation function, such as sigmoid or tanh, $g(\cdot)$ depends on our task, $W \in \mathbb{R}^{2d \times d}$ is the weight parameter for projecting child input vectors $x_1, x_2, x_3 \in \mathbb{R}^{d}$ into the parent vector, $W_y$ is a weight parameter for computing output vector, and $b, b_y$ are biases. If we want to train RecNN for classification tasks, $g(\cdot)$ can be defined as a softmax function. 
\begin{figure}[h]
    \centering
    \includegraphics[width=8.8cm]{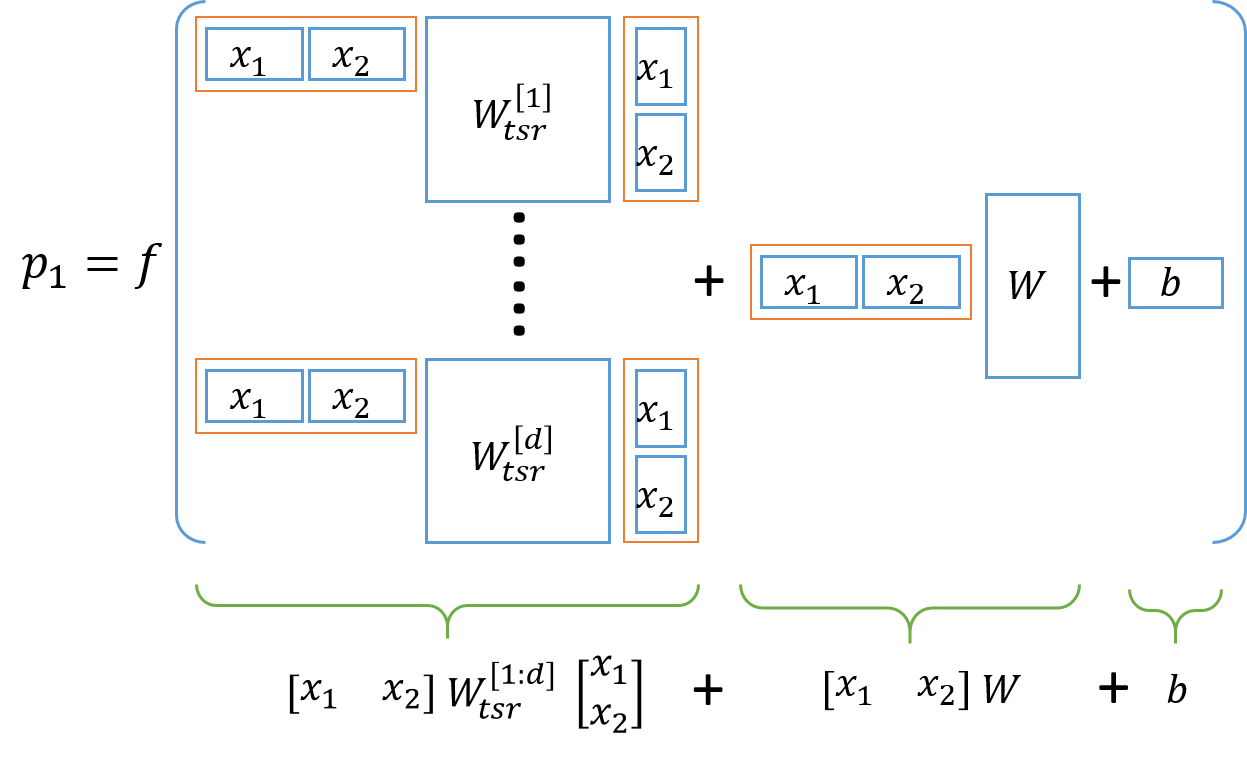}
    \caption{Calculating vector $p_1$ from left input $x_1$ and right input $x_2$ based on Eq. \ref{eq:rntnp1}}
    \label{fig:rntn}
\end{figure}
However, standard RecNNs have several limitations, where two vectors only implicitly interact with addition before applying a nonlinear activation function on them \cite{socher2013recursive} and standard RecNNs are not able to model very long-term dependency on tree structures. Zhu et al.\cite{zhu2015long} proposed the gating mechanism into standard RecNN model to solve the latter problem. For the former limitation, the RecNN performance can be improved by adding more interaction between the two input vectors. Therefore, a new architecture called a Recursive Neural Tensor Network (RecNTN) tried to overcome the previous problem by adding interaction between two vectors using a tensor product, which is connected by tensor weight parameters. Each slice of the tensor weight can be used to capture the specific pattern between the left and right child vectors. For RecNTN, value $p_1$ from Eq. \ref{eq:p1} and \ref{eq:p2} is defined by:
\begin{eqnarray}
p_1 &=& f\left(
    \begin{bmatrix} x_1 & x_2 \end{bmatrix} W_{tsr}^{[1:d]} \begin{bmatrix} x_1 \\ x_2 \end{bmatrix} + \begin{bmatrix} x_1 & x_2 \end{bmatrix} W + b \right) \label{eq:rntnp1}\\ 
p_2 &=& f\left(
    \begin{bmatrix} p_1 & x_3 \end{bmatrix} W_{tsr}^{[1:d]} \begin{bmatrix} p_1 \\ x_3 \end{bmatrix} + \begin{bmatrix} p_1 & x_3 \end{bmatrix} W + b \right)
\end{eqnarray} where $W_{tsr}^{[1:d]} \in \mathbb{R}^{2d \times 2d \times d}$ is the tensor weight to map the tensor product between two children vectors. Each slice $W_{tsr}^{[i]}$ is a matrix $\mathbb{R}^{2d \times 2d}$. For more details, we visualize the calculation for $p_1$ in Fig. \ref{fig:rntn}.

\section{Proposed Architecture} \label{sec:proposed}
\subsection{Gated Recurrent Unit Recurrent Neural Tensor Network (GRURNTN)}
\begin{figure}[h]
    \centering
    \includegraphics[width=8.8cm]{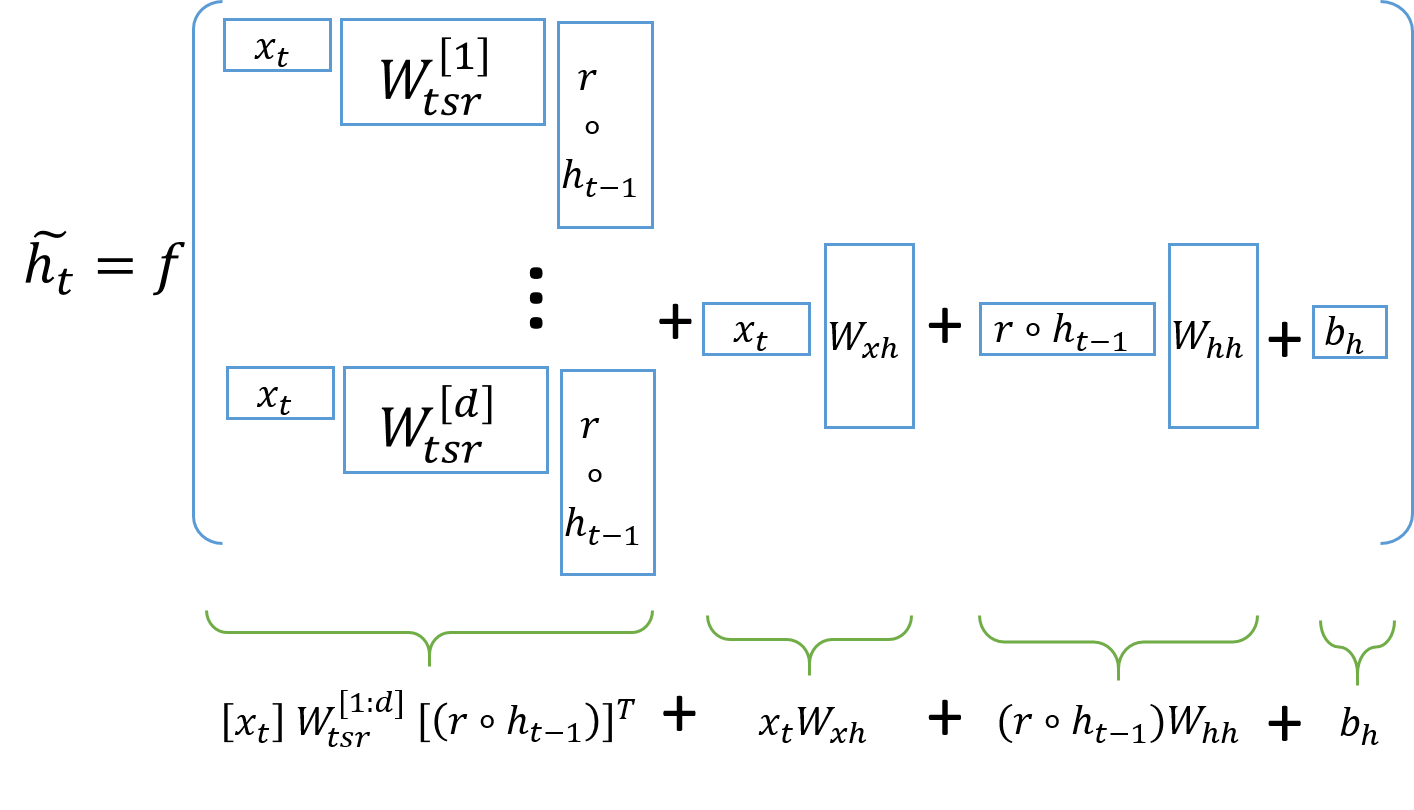}
    \caption{Calculating candidate hidden layer $\tilde{h_t}$ from current input $x_t$ and previous hidden layer multiplied by reset gate $r \cdot h_{t_1}$ based on Eq. \ref{eq:grurntn}}
    \label{fig:grurntn}
\end{figure}

Previously in Sections \ref{sec:gatedrnn} and \ref{sec:recntn}, we discussed that the gating mechanism concept can helps RNNs learn long-term dependencies from sequential input data and that adding more powerful interaction between the input and hidden layers simultaneously with the tensor product operation in a bilinear form improves neural network performance and expressiveness. By using tensor product, we increase our model expressiveness by using second-degree polynomial interactions, compared to first-degree polynomial interactions on standard dot product followed by addition in common RNNs architecture. Therefore, in this paper we proposed a Gated Recurrent Neural Tensor Network (GRURNTN) to combine these two advantages into an RNN architecture. In this architecture, the tensor product operation is applied between the current input and previous hidden layer multiplied by the reset gates for calculating the current candidate hidden layer values. The calculation is parameterized by tensor weight. To construct a GRURNTN, we defined the formulation as:
\begin{eqnarray}
r_t &=& \sigma(x_t W_{xr} + h_{t-1} W_{hr} + b_r) \nonumber \\
z_t &=& \sigma(x_t W_{xz} + h_{t-1} W_{hz} + b_z) \nonumber \\
\tilde{h_t} &=& f\left( \begin{bmatrix} x_t & (r \odot h_{t-1}) \end{bmatrix} W_{tsr}^{[1:d]} \begin{bmatrix} x_t \\ (r \odot h_{t-1}) \end{bmatrix} \right. \nonumber \\ 
& & \left. + x_t W_{xh} + (r_t \odot h_{t-1}) W_{hh} + b_h \right) \\
h_t &=& (1 - z_t) \odot h_{t-1} + z_t \odot \tilde{h_t} \nonumber 
\end{eqnarray}
where $W_{tsr}^{[1:d]} \in \mathbb{R}^{(i+d) \times (i+d) \times d}$ is a tensor weight for mapping the tensor product between the input-hidden layer, $i$ is the input layer size, and $d$ is the hidden layer size. Alternatively, in this paper we use a simpler bilinear form for calculating $\tilde{h_t}$:
\begin{eqnarray}
\tilde{h_t} &=& f\left( \begin{bmatrix} x_t \end{bmatrix} W_{tsr}^{[1:d]} \begin{bmatrix} (r_t \odot h_{t-1}) \end{bmatrix}^{\intercal} \right. \nonumber \\ 
 & & \left. + x_t W_{xh} + (r_t \odot h_{t-1}) W_{hh} + b_h \right) \label{eq:grurntn}
\end{eqnarray} where $W_{tsr}^{[i:d]} \in \mathbb{R}^{i \times d \times d}$ is a tensor weight. Each slice $W_{tsr}^{[i]}$ is a matrix $\mathbb{R}^{i \times d}$. The advantage of this asymmetric version is that we can still maintain the interaction between the input and hidden layers through a bilinear form. We reduce the number of parameters from the original neural tensor network formulation by using this asymmetric version. Fig. \ref{fig:grurntn} visualizes the $\tilde{h_t}$ calculation in more detail.

\subsection{LSTM Recurrent Neural Tensor Network (LSTMRNTN)}
\begin{figure}[h]
    \centering
    \includegraphics[width=8.8cm]{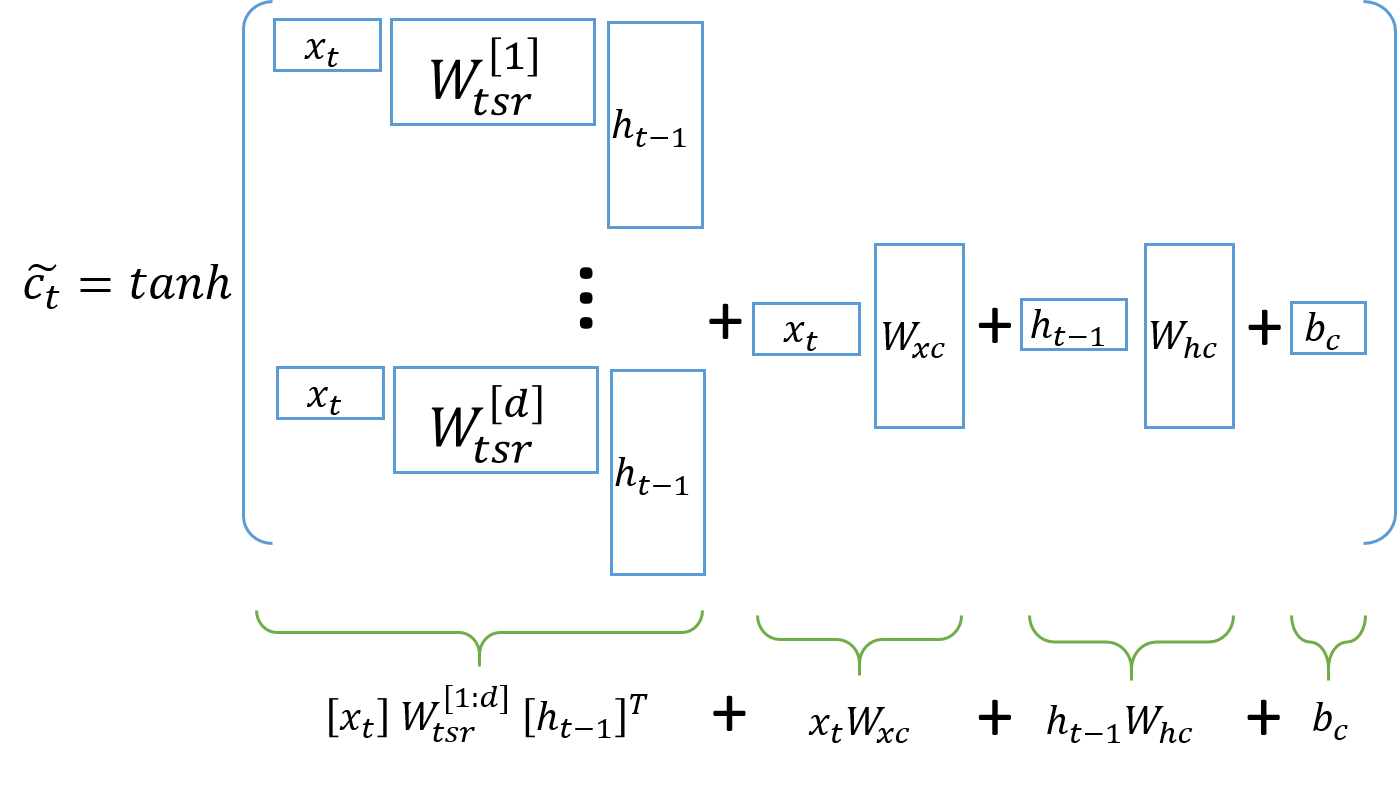}
    \caption{Calculating candidate cell layer $\tilde{c_t}$ from current input $x_t$ and previous hidden layer $h_{t-1}$ based on Eq. \ref{eq:lstmrntn}}
    \label{fig:lstmrntn}
\end{figure}

As with GRURNTN, we also applied the tensor product operation for the LSTM unit to improve its performance. In this architecture, the tensor product operation is applied between the current input and the previous hidden layers to calculate the current memory cell. The calculation is parameterized by the tensor weight. We call this architecture a Long Short Term Memory Recurrent Neural Tensor Network (LSTMRNTN). To construct an LSTMRNTN, we defined its formulation:
\begin{eqnarray}
i_t &=& \sigma(x_t W_{xi} + h_{t-1} W_{hi} + c_{t-1} W_{ci} + b_i) \nonumber \\
f_t &=& \sigma(x_t W_{xf} + h_{t-1} W_{hf} + c_{t-1} W_{cf} + b_f) \nonumber \\
\tilde{c_t} &=& \tanh \left( \begin{bmatrix} x_t \end{bmatrix} W_{tsr}^{[1:d]} \begin{bmatrix} h_{t-1} \end{bmatrix}  \right. \nonumber \\
& & \left. + x_t W_{xc} + h_{t-1} W_{hc} + b_c \right) \label{eq:lstmrntn} \\
c_t &=& f_t \odot c_{t-1} + i_t \odot \tilde{c_t} \\
o_t &=& \sigma(x_t W_{xo} + h_{t-1} W_{ho} + c_t W_{co} + b_o) \nonumber \\
h_t &=& o_t \odot \tanh(c_t) \nonumber
\end{eqnarray}
where $W_{tsr}^{[1:d]} \in R^{i \times d \times d}$ is a tensor weight to map the tensor product between current input $x_t$ and previous hidden layer $h_{t-1}$ into our candidate cell $\tilde{c_t}$. Each slice $W_{tsr}^{[i]}$ is a matrix $\mathbb{R}^{i \times d}$. Fig. \ref{fig:lstmrntn} visualizes the $\tilde{c_t}$ calculation in more detail.
\subsection{Optimizing Tensor Weight using Backpropagation Through Time}
In this section, we explain how to train the tensor weight for our proposed architecture. Generally, we use backpropagation to train most neural network models \cite{rumelhart1986learning}. For training an RNN, researchers tend to use backpropagation through time (BPTT) where the recurrent operation is unfolded as a feedforward neural network along with the time-step when we backpropagate the error \cite{werbos1990backpropagation, mikolov2012statistical}. Sometimes we face a performance issue when we unfold our RNN on such very long sequences. To handle that issue, we can use the truncated BPTT \cite{mikolov2010recurrent} to limit the number of time-steps when we unfold our RNN during backpropagation.

Assume we want to do segment classification \cite{graves2012supervised} with an RNN trained as function $f : \mathbf{x} \rightarrow \mathbf{y}$, where $\mathbf{x} = (x_1,...,x_T)$ as an input sequence and $\mathbf{y} = (y_1,...,y_T)$ is an output label sequence. In this case, probability output label sequence $y$, given input sequence $\mathbf{x}$, is defined as:
\begin{eqnarray}
P(\mathbf{y}|\mathbf{x}) = \prod_{i=1}^{T}P(y_i | x_1,..,x_i)
\end{eqnarray} 
Usually, we transform likelihood $P(\mathbf{y}|\mathbf{x})$ into a negative log-likelihood:
\begin{eqnarray}
E(\theta) &=& -\log P(\mathbf{y}|\mathbf{x}) = -\log \left(\prod_{i=1}^{T} P(y_{i}|x_1,..,x_i)\right) \\
 &=& -\sum_{i=1}^{T} \log P(y_i | x_1,..,x_i)
\end{eqnarray} and our objective is to minimize the negative log-likelihood w.r.t all weight parameters 
$\theta$.
To optimize $W_{tsr}^{[1:d]}$ weight parameters, we need to find derivative $E(\theta)$ w.r.t $W_{tsr}^{[1:d]}$ : 
\begin{eqnarray}
\frac{\partial E(\theta)}{\partial W_{tsr}^{[1:d]}} &=& \sum_{i=1}^{T} \frac{\partial E_i(\theta)}{\partial W_{tsr}^{[1:d]}} \nonumber
\end{eqnarray}
For applying backpropagation through time, we need to unfold our GRURNTN and backpropagate the error from $E_i(\theta)$ to all candidate hidden layer $\tilde{h_j}$ to accumulate $W_{tsr}^{[1..d]}$ gradient where $j \in [1..i]$. If we want to use the truncated BPTT to ignore the history past over $K$ time-steps, we can limit $j \in [max(1, i-K) .. i]$. We define the standard BPTT on GRURNTN to calculate $\partial E_i(\theta) / \partial W_{tsr}^{[1..d]}$:
\begin{eqnarray}
\frac{\partial E_i(\theta)}{\partial W_{tsr}^{[1:d]}} &=& \sum_{j=1}^{i} \frac{\partial E_i(\theta)}{\partial \tilde{h_j}} \frac{\partial \tilde{h_j}}{\partial W_{tsr}^{[1:d]}} \nonumber \\
&=& \sum_{j=1}^{i} \frac{\partial E_i(\theta)}{\partial \tilde{h_j}}\frac{\partial \tilde{h_j}}{\partial a_j} \frac{\partial a_j}{\partial W_{tsr}^{[1:d]}} \nonumber \\
&=& \sum_{j=1}^{i} \frac{\partial E_i(\theta)}{\partial \tilde{h_j}} f'(a_j) \begin{bmatrix} x_j \end{bmatrix}^{\intercal} \begin{bmatrix} (r_j \odot h_{j-1}) \end{bmatrix} \label{eq:derivgrurntn}
\end{eqnarray} 
where \begin{eqnarray} a_j &=& \left( \begin{bmatrix} x_j \end{bmatrix} W_{tsr}^{[1:d]} \begin{bmatrix} (r_j \odot h_{j-1}) \end{bmatrix}^{\intercal} \right. \nonumber \\ & & \left.  + x_j W_{xh} + (r_j \odot h_{j-1}) W_{hh} + b_h \right) \nonumber 
\end{eqnarray} 
and $f'(\cdot)$ is a function derivative from our activation function :  
\begin{eqnarray}
f'(a_j) =
\begin{cases}
 (1-f(a_j)^2), & \text{if } f(\cdot) \text{ is $\tanh$ function} \\
 f(a_j)(1-f(a_j)), & \text{if } f(\cdot) \text{ is sigmoid function}
\end{cases} \nonumber 
\end{eqnarray}
For LSTMRNTN, we also need to unfold our LSTMRNN and backpropagate the error from $E_i(\theta)$ to all cell layers $c_j$ to accumulate $W_{tsr}^{[1..d]}$ gradients where $j \in [1..i]$. We define the standard BPTT on LSTMRNTN to calculate $\partial E_i(\theta) / \partial W_{tsr}^{[1..d]}$:
\begin{eqnarray}
\frac{\partial E_i(\theta)}{\partial W_{tsr}^{[1:d]}} &=& \sum_{j=1}^{i} \frac{\partial E_i{(\theta)}}{\partial c_j} \frac{\partial c_j}{\partial W_{tsr}^{[1:d]}} \nonumber \\
& = & \sum_{j=1}^{i} \frac{\partial E_i{(\theta)}}{\partial c_j} \frac{\partial c_j}{\partial \tanh(a_j)} \frac{\partial \tanh(a_j)} {\partial a_j} \frac{\partial a_j}{\partial W_{tsr}^{[1:d]}} \nonumber \\
& = & \sum_{j=1}^{i} \frac{\partial E_i{(\theta)}}{\partial c_j} i_j (1-\tanh^2(a_j)) \begin{bmatrix} x_j \end{bmatrix}^{\intercal} \begin{bmatrix} h_{j-1} \end{bmatrix} \label{eq:derivlstmrntn} 
\end{eqnarray} 
where \begin{eqnarray} a_j &=& \left(\begin{bmatrix} x_j \end{bmatrix} W_{tsr}^{[1:d]} \begin{bmatrix} h_{j-1} \end{bmatrix} + x_j W_{xc} + h_{j-1} W_{hc} + b_c \right)  \end{eqnarray}. 
In both proposed models, we can see partial derivative ${\partial E_i(\theta)} / {\partial W_{tsr}^{[1:d]}}$ in Eqs. \ref{eq:derivgrurntn} and \ref{eq:derivlstmrntn}, the derivative from the tensor product w.r.t the tensor weight parameters depends on the values of our input and hidden layers. Then all the slices of tensor weight derivative are multiplied by the error from their corresponding pre-activated hidden unit values. From these derivations, we are able to see where each slice of tensor weight is learned more directly from their input and hidden layer values compared by using standard addition operations.
After we accumulated every parameter's gradients from all the previous time-steps, we use a stochastic gradient optimization method such as AdaGrad \cite{duchi2011adaptive} to optimize our model parameters.

\section{Experiment Settings} \label{sec:expr}
Next we evaluate our proposed GRURNTN and LSTMRNTN models against baselines GRURNN and LSTMRNN with two different tasks and datasets.
\subsection{Datasets and Tasks}
We used a PennTreeBank (PTB) corpus\footnote{\url{https://www.cis.upenn.edu/~treebank/}}, which is a standard benchmark corpus for statistical language modeling. A PTB corpus is a subset of the WSJ corpus. In this experiment, we followed the standard preprocessing step that was done by previous research \cite{mikolov2012statistical}. The PTB dataset is divided as follows: a training set from sections 0-20 with total 930.000 words, a validation set from sections 21-22 with total 74.000 words, and a test set from sections 23-24 with total 82.000 words. The vocabulary is limited to the 10.000 most common words, and all words outside are mapped into a "$<$unk$>$" token.  We used the preprocessed PTB corpus from the RNNLM-toolkit website\footnote{\url{http://www.rnnlm.org/}}. 

We did two different language modeling tasks. First, we experimented on a word-level language model where our RNN predicts the next word probability given the previous words and current word. We used perplexity (PPL) to measure our RNN performance for word-level language modeling. The formula for calculating the PPL of word sequence $X$ is defined by:
\begin{eqnarray}
PPL = 2^{-\frac{1}{N}\sum_{i=1}^{N} \log_2 P(X_i|X_{1..{i-1}})} 
\end{eqnarray}
Second, we experimented on a character-level language model where our RNN predicts the next character probability given the previous characters and current character. We used the average number of bits-per-character (BPC) to measure our RNN performance for character-level language modeling. The formula for calculating the BPC of character sequence $X$ is defined by: 
\begin{eqnarray} 
BPC = -\frac{1}{N}\left(\sum_{i=1}^{N}\log_2{p(X_i|X_{1..{i-1}})} \right)
\end{eqnarray} 

\subsection{Experiment Models}
In this experiment, we compared the performance from our baseline models GRURNN and LSTMRNN with our proposed GRURNTN and LSTMRNTN models. We used the same dimensions for the embedding matrix to represent the words and characters as the vectors of real numbers.

For the word-level language modeling task, we used 256 hidden units for GRURNTN and LSTMRNTN, 860 for GRURNN, and 740 for LSTMRNN. All of these models use 128 dimensions for word embedding. We used dropout regularization with $p=0.5$ dropout probability for GRURNTN and LSTMRNTN and $p=0.6$ for our baseline model. The total number of free parameters for GRURNN and GRURNTN were about 12 million and about 13 million for LSTMRNN and LSTMRNTN.

For the character-level language modeling task, we used 256 hidden units for GRURNTN and LSTMRNTN, 820 for GRURNN, and 600 for LSTMRNTN. All of these models used 32 dimensions for character embedding. We used dropout regularization with $p=0.25$ dropout probability. The total number of free parameters for GRURNN and GRURNTN was about 2.2 million and about 2.6 million for LSTMRNN and LSTMRNTN.

We constrained our baseline GRURNN to have a similar number of parameters as the GRURNTN model for a fair comparison. We also applied such constraints on our baseline LSTMRNN to LSTMRNTN model.

For all the experiment scenarios, we used AdaGrad for our stochastic gradient optimization method with mini-batch training and a batch size of 15 sentences. We multiplied our learning rate with a decay factor of 0.5 when the cost from the development set for current epoch is greater than previous epoch. We also used a rescaling trick on the gradient \cite{pascanu2012difficulty} when the norm was larger than 5 to avoid the issue of exploding gradients. For initializing the parameters, we used the orthogonal weight initialization trick \cite{saxe2013exact} on every model. 

\section{Results and Analysis} \label{sec:results}
\subsection{Character-level Language Modeling}
\begin{figure}[h]
    \centering
    \includegraphics[width=8.7cm]{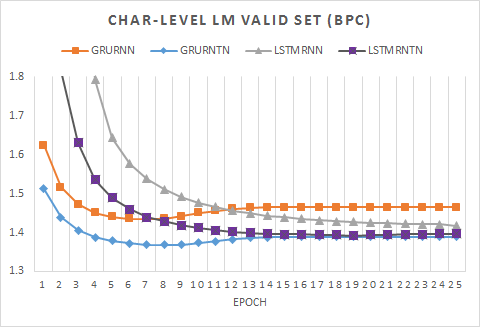}
    \caption{Comparison among GRURNN, GRURNTN, LSTMRNN, and LSTMRNTN bits-per-character (BPC) per epoch on PTB validation set. Note : a lower BPC score is better}
    \label{fig:reportchar}
\end{figure}

\begin{table}[]
\centering
\caption{PennTreeBank test set BPC}
\label{tbl:reportchar}
\begin{tabular}{|c|c|}
\hline
\textbf{Model} & \textbf{Test BPC} \\ \hline
NNLM \cite{mikolov2012subword, bengio2003neural} & 1.57 \\ \hline
BPTT-RNN \cite{mikolov2012subword} & 1.42 \\ \hline
HF-MRNN \cite{mikolov2012subword, sutskever2011generating} & 1.41 \\ \hline
sRNN \cite{pascanu2013construct}       & 1.41            \\ \hline
DOT(S)-RNN \cite{pascanu2013construct}       & 1.39            \\ \hline
LSTMRNN (w/ adapt. noise, w/o dyn. eval) \footnote{\label{note:adaptnoise}Adaptive noise regularization is where the noise variance is learned and applied with the weight} \footnote{\label{note:dyneval}Dynamic evaluation approach updates the model parameter during processing on the test data (only updated once per test dataset). Our baseline and proposed model experiment did not use dynamic evaluation.} \cite{graves2013generating} & 1.26 \\ \hline
LSTMRNN (w/ adapt. noise, w/ dyn. eval) \footnotemark[\getrefnumber{note:adaptnoise}] \footnotemark[\getrefnumber{note:dyneval}] \cite{graves2013generating} & 1.24 \\ \hline

GRURNN (our baseline)        & 1.39            \\ \hline
LSTMRNN (our baseline)       & 1.37            \\ \hline
GRURNTN (proposed) & 1.33            \\ \hline
LSTMRNTN (proposed) & 1.34            \\ \hline
\end{tabular}
\end{table}
In this section, we report our experiment results on PTB character-level language modeling using our baseline models GRURNN and LSTMRNN as well as our proposed models GRURNTN and LSTMRNTN. Fig. \ref{fig:reportchar} shows performance comparisons from every model based on the validation set's BPC per epoch. In this experiment, GRURNN made faster progress than LSTMRNN, but eventually LSTMRNN converged into a better BPC based on the development set. Our proposed model GRURNTN made faster and quicker progress than LSTMRNTN and converged into a similar BPC in the last epoch. Both proposed models produced lower BPC than our baseline models from the first epoch to the last epoch.

Table \ref{tbl:reportchar} shows PTB test set BPC among our baseline models, our proposed models and several published results.
Our proposed model GRURNTN and LSTMRNTN outperformed both baseline models. GRURNTN reduced the BPC from 1.39 to 1.33 (0.06 absolute / 4.32\% relative BPC) from the baseline GRURNN, and LSTMRNTN reduced the BPC from 1.37 to 1.34 (0.03 absolute / 2.22\% relative BPC) from the baseline LSTMRNN. Overall, GRURNTN slightly outperformed LSTMRNTN, and both proposed models outperformed all of the baseline models on the character-level language modeling task.

\subsection{Word-level Language Modeling}

\begin{figure}[h]
    \centering
    \includegraphics[width=8.7cm]{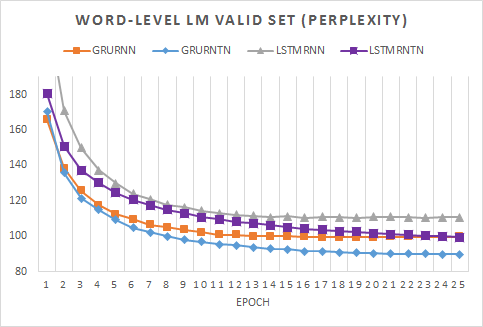}
    \caption{Comparison among GRURNN, GRURNTN, LSTMRNN and LSTMRNTN perplexity (PPL) per epoch on the PTB validation set. Note : a lower PPL value is better}
    \label{fig:reportword}
\end{figure}
In this section, we report our experiment results on PTB word-level language modeling using our baseline models GRURNN and LSTMRNN and our proposed models GRURNTN and LSTMRNTN. Fig. \ref{fig:reportword} compares the performance from every models based on the validation set's PPL per epoch. In this experiment, GRURNN made faster progress than LSTMRNN. Our proposed GRURNTN's progress was also better than LSTMRNTN. The best model in this task was GRURNTN, which had a consistently lower PPL than the other models.

\begin{table}[]
\centering
\caption{PennTreeBank test set PPL}
\label{tbl:reportword}
\begin{tabular}{|c|c|}
\hline
\textbf{Model} & \textbf{Test PPL} \\ \hline
N-Gram \cite{mikolov2012statistical}       & 141              \\ \hline
RNNLM (w/o dyn. eval) \footnotemark[\getrefnumber{note:dyneval}] \cite{mikolov2012statistical} & 124.7              \\ \hline
RNNLM (w/ dyn. eval) \footnotemark[\getrefnumber{note:dyneval}] \cite{mikolov2012statistical} & 123.2              \\ \hline
SCRNN \cite{mikolov2014learning}        & 115              \\ \hline
sRNN \cite{pascanu2013construct}        & 110.0              \\ \hline
DOT(S)-RNN \cite{pascanu2013construct}        & 107.5              \\ \hline
GRURNN (our baseline)        & 97.78              \\ \hline
LSTMRNN (our baseline)        & 108.26
            \\ \hline
GRURNTN (proposed)       & 87.38             \\ \hline
LSTMRNTN (proposed)   & 96.97             \\ \hline
\end{tabular}
\end{table}
Table \ref{tbl:reportchar} shows the PTB test set PPL among our baseline models, proposed models, and several published results. 
Both our proposed models outperformed their baseline models. GRURNTN reduced the perplexity from 97.78 to 87.38 (10.4 absolute / 10.63\% relative PPL) over the baseline GRURNN and LSTMRNTN reduced the perplexity from 108.26 to 96.97 (11.29 absolute / 10.42\% relative PPL) over the baseline LSTMRNN. Overall, LSTMRNTN improved the LSTMRNN model and its performance closely resembles the baseline GRURNN. However, GRURNTN outperformed all the baseline models as well as the other models by a large margin.
\section{Related Work} \label{sec:rel}
Representing hidden states with deeper operations was introduced just a few years ago \cite{pascanu2013construct}. In these works, Pascanu et al.\cite{pascanu2013construct} use additional nonlinear layers for representing the transition from input to hidden layers, hidden to hidden layers, and hidden to output layers. They also improved the RNN architecture by a adding shortcut connection in the deep transition by skipping the intermediate layers. Another work from \cite{chung2015gated} proposed a new RNN design for a stacked RNN model called Gated Feedback RNN (GFRNN), which adds more connections from all the previous time-step stacked hidden layers into the current hidden layer computations. 
Despite adding additional transition layers and connection weight from previous hidden layers, all of these models still represent the input and hidden layer relationships by using linear projection, addition and nonlinearity transformation. 

On the tensor-based models, Irsoy et al.\cite{irsoy2014modeling} proposed a simple RNN with a tensor product between the input and hidden layers. Such architecture resembles RecNTN, given a parse tree with a completely unbalanced tree on one side. Another work from \cite{hutchinson2013tensor} also use tensor products for representing hidden layers on DNN. By splitting the weight matrix into two parallel weight matrices, they calculated two parallel hidden layers and combined the pair of hidden layers using a tensor product. However, since not all of those models use a gating mechanism, the tensor parameters and tensor product operation can not be fully utilized because of the vanishing (or exploding) gradient problem. 

On the recurrent neural network-based model, Sutskever et al.\cite{sutskever2011generating} proposed multiplicative RNN (mRNN) for character-level language modeling using tensor as the weight parameters. They proposed two different models. The first selected a slice of tensor weight based on the current character input, and the second improved the first model with factorization for constructing a hidden-to-hidden layer weight. However, those models fail to fully utilize the tensor weight with the tensor product. After they selected the weight matrix based on the current input information, they continue to use linear projection, addition, and nonlinearity for interacting between the input and hidden layers.

To the best of our knowledge, none of these works combined the gating mechanism and tensor product concepts into a single neural network architecture. In this paper, we built a new RNN by combining gating units and tensor products into a single RNN architecture. We expect that our proposed GRURNTN and LSTMRNTN architecture will improve the RNN performance for modeling temporal and sequential datasets.

\section{Conclusion} \label{sec:conclusion}
We presented a new RNN architecture by combining the gating mechanism and tensor product concepts. Our proposed architecture can learn long-term dependencies from temporal and sequential data using gating units as well as more powerful interaction between the current input and previous hidden layers by introducing tensor product operations. From our experiment on the PennTreeBank corpus, our proposed models outperformed the baseline models with a similar number of parameters in character-level language modeling and word-level language modeling tasks. In a character-level language modeling task, GRURNTN obtained 0.06 absolute (4.32\% relative) BPC reduction over GRURNN, and LSTMRNTN obtained 0.03 absolute (2.22\% relative) BPC reduction over LSTMRNN. In a word-level language modeling task, GRURNTN obtained 10.4 absolute (10.63\% relative) PPL reduction over GRURNN, and LSTMRNTN obtained 11.29 absolute (10.42\% relative PPL) reduction over LSTMRNN. In the future, we will investigate the possibility of combining our model with other stacked RNNs architecture, such as Gated Feedback RNN (GFRNN). We would also like to explore other possible tensor operations and integrate them with our RNN architecture. By applying these ideas together, we expect to gain further performance improvement. Last, for further investigation we will apply our proposed models to other temporal and sequential tasks, such as speech recognition and video recognition.

\section{Acknowledgements}
Part of this research was supported by JSPS KAKENHI Grant Number 26870371.

\bibliographystyle{IEEEtran}
\bibliography{refs}

\end{document}